\documentclass[sigconf]{acmart}
\usepackage{balance}
\usepackage{calc}
\usepackage[table]{xcolor}
\usepackage{booktabs}
\usepackage{multirow}
\usepackage{graphicx}
\AtBeginDocument{%
  }

\copyrightyear{2026}
\acmYear{2026}
\setcopyright{cc}
\setcctype{by}
\acmConference[SIGIR '26]{Proceedings of the 49th International ACM SIGIR Conference on Research and Development in Information Retrieval}{July 20--24, 2026}{Melbourne, VIC, Australia}
\acmBooktitle{Proceedings of the 49th International ACM SIGIR Conference on Research and Development in Information Retrieval (SIGIR '26), July 20--24, 2026, Melbourne, VIC, Australia}
\acmDOI{10.1145/3805712.3809583}
\acmISBN{979-8-4007-2599-9/2026/07}

\begin{document}

\title{MECI: Multi-Element Collaborative Interaction for Multimodal Entity Linking}


\author{Jie Peng}
\email{pengjie@nudt.edu.cn}
\affiliation{
  \institution{College of Computer Science and Technology, National University of Defense Technology}
  \city{Changsha}
  \state{Hunan}
  \country{China}
}

\author{Yongxue Shan}
\email{shanyongxue001@nudt.edu.cn}
\affiliation{%
  \institution{College of Computer Science and Technology, National University of Defense Technology}
  \city{Changsha}
  \state{Hunan}
  \country{China}
}

\author{Yongfu Zha}
\email{zhayongfu@nudt.edu.cn}
\affiliation{%
  \institution{College of Computer Science and Technology, National University of Defense Technology}
  \city{Changsha}
  \state{Hunan}
  \country{China}
}

\author{Xiaodong Wang}
\authornote{Corresponding author.}  
\email{xdwang@nudt.edu.cn}
\affiliation{%
  \institution{College of Computer Science and Technology, National University of Defense Technology}
  \city{Changsha}
  \state{Hunan}
  \country{China}
}

\renewcommand{\shortauthors}{Jie Peng, Yongxue Shan, Yongfu Zha, and Xiaodong Wang}

\begin{abstract}
Multimodal Entity Linking (MEL) aims to disambiguate mentions in multimodal contexts by grounding them to specific entities in a knowledge base. A pivotal challenge in MEL is capturing multi-level correspondences: the semantic consistency between mention-entity pairs and the complementary correlations across modalities. However, existing methods often suffer from element dominance due to their reliance on coupled interactions or coarse global aggregations. In response, we propose the \textbf{M}ulti-\textbf{E}lement \textbf{C}ollaborative \textbf{I}nteraction (\textbf{MECI}) framework. First, to capture multi-element mention-entity correspondences, we develop a Multi-view Experts Network that leverages a “divide-and-conquer” strategy for decoupled feature learning to mitigate element dominance, supported by a KL-guided routing mechanism that governs expert specialization and collaboration. Furthermore, to model cross-modal complementary correlations, we propose a Hierarchical Multimodal Interaction Module, where a dynamic modality-aware weighting network refines interactions across hierarchical semantic levels, thereby integrating multi-granular evidence to counteract element dominance. Finally, we incorporate a generative semantic refinement stage that utilizes large language models for zero-shot re-ranking. Extensive experiments on WikiDiverse, RichpediaMEL, and WikiMEL show that MECI consistently outperforms state-of-the-art baselines, improving Hits@1 by 1.95\%, 7.30\%, and 2.31\%, respectively.
\end{abstract}




\maketitle

\section{Introduction}
\begin{figure}[t]
    \centering
    \includegraphics[width=\linewidth]{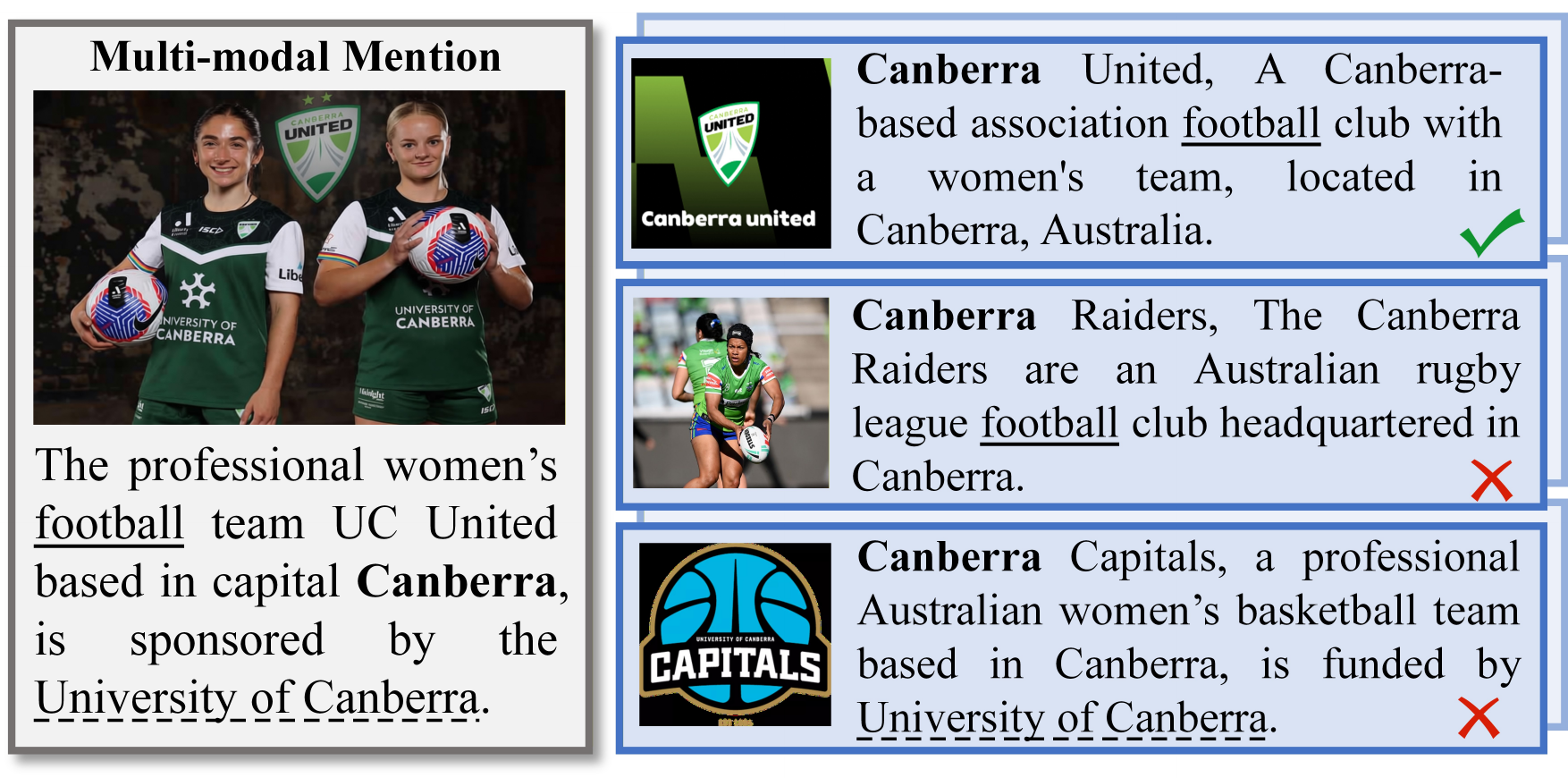}
    \caption{An illustration of the MEL task: multimodal mentions (left) and candidate entities (right).}
    \label{fig1}
\end{figure}

Entity Linking (EL) serves as a fundamental bridge connecting real-world information to knowledge bases. Over the past years, large efforts focused on text-based EL~\cite{ref1,ref2}. However, images and text have become the most widely seen medium for publishing and understanding web information. Therefore, Multimodal Entity Linking (MEL)~\cite{ref3,ref4} has emerged as a pivotal evolution, attracting attention from both academia and industry. By leveraging textual and visual cues, MEL plays a crucial role in knowledge-intensive applications such as semantic search~\cite{ref50, ref5-1}, question answering~\cite{ref6, ref7, ref8}, and dialog systems~\cite{ref9, ref10, ref11}. As illustrated in Figure \ref{fig1}, given a multimodal mention, the MEL model aims to link it to the entity \texttt{Canberra United} in a multimodal knowledge base.

To effectively leverage multimodal information in MEL, a pivotal challenge is to identify and utilize the multi-element correspondences inherent in entity and mention contexts: \textbf{1) Multi-element correspondences between mentions and entities}: Both the mention context and the entity representation often contain multiple consistent cues that facilitate semantic matching. As shown in Figure \ref{fig1}, the mention and the entity share textual elements such as \texttt{Canberra}, \texttt{football}, and \texttt{University of Canberra}. These correspondences highlight semantic similarities and provide explicit evidence for accurate entity linking, thereby reducing ambiguity and enhancing matching confidence. \textbf{2) Multi-element correspondences between modalities}: The textual content and visual patches may exhibit correlative correspondences that are pivotal for comprehensive understanding. For instance, in Figure \ref{fig1}, the presence of a football in the image, combined with the textual context, helps definitively identify the entity as a football team—rather than a rugby or basketball team. Such cross-modal interactions allow textual and visual information to complement each other, enriching the model’s discriminative capacity for entity disambiguation. 

To facilitate these multi-element correspondences, existing studies propose sophisticated correlative mechanisms, such as hierarchical gating mechanisms~\cite{ref12,ref13,ref14}, attention-based paradigms~\cite{ref15,ref16,ref17}, graph-based aggregation schemes~\cite{ref19,ref20,ref21}, and adaptive fusion strategies~\cite{ref22,ref23,ref24}. 
Despite their progress, these methods face inherent challenges in handling the complexity of heterogeneous inputs. Specifically, hierarchical gating and attention-based approaches often employ coupled interaction designs across element levels (e.g., coupling coarse- and fine-grained semantics), which hinders adaptive adjustment to the characteristics of input semantics and readily leads to element dominance issues in the competition among heterogeneous features. Furthermore, graph-based and fusion approaches typically compress textual and visual features into a coarse global representation. This representation-level bottleneck makes it difficult to capture fine-grained, context-dependent associations, such as fine-grained visual correlations between entity and mention. In summary, \textbf{existing designs tend to over-rely on prominent elements, leading to element dominance}, where stronger modalities often overshadow weaker ones (modality dominance) or coarse global representations suppress fine-grained details (granularity bias). Such dependencies severely limit their capacity to model multi-element correspondences comprehensively, ultimately compromising entity linking accuracy in complex multimodal scenarios.

To mitigate these limitations and comprehensively capture multi-element correspondences, we propose the \textbf{Multi-Element Collaborative Interaction (MECI)} framework for multimodal entity linking. First, targeting the multi-element correspondences between mentions and entities, we introduce the \textbf{Multi-view Experts Network (MVEN)}, which employs a ``divide-and-conquer'' strategy to capture both universal and view-specific characteristics. To regulate expert specialization and knowledge sharing, we propose a KL-guided routing mechanism that utilizes structural priors to adaptively coordinate information flow. Then, incorporating with a hierarchical matching strategy, MVEN provides specialized processing capacities for multi-element correlation analysis, effectively mitigating the element dominance problem. However, while MVEN effectively refines view-specific representations, these features remain relatively isolated. To bridge this gap and fully exploit the multi-element correspondences between modalities, we develop the \textbf{Hierarchical Multimodal Interaction Module (HMIM)}. This module employs a dynamic cross-modal weighting network to refine interactions at distinct granularity levels—avoiding the granularity bias inherent in unified interaction paradigms—and utilizes a granularity-aware gating network to dynamically integrate multi-granular evidence for comprehensive cross-modal understanding. Finally, to further enhance linking precision, we introduce a finetuning-free generative semantic refinement stage, which reformulates candidate re-ranking as a zero-shot multiple-choice task to leverage the parametric knowledge of frozen Large Language Models (LLMs). In summary, our main contributions are threefold:
\enlargethispage{0.01\baselineskip}
\begin{itemize}
    \item We propose the Multi-view Experts Network for mention-entity correspondences. It employs a ``divide-and-conquer'' strategy with specialized routing and matching to decouple heterogeneous features, mitigating element dominance.
    
    \item We develop a Hierarchical Multimodal Interaction Module for cross-modal correspondences. It employs dynamic weighting to avoid granularity bias and a gating network to incorporate multi-granular evidence.
    
    \item Experiments on WikiMEL, RichpediaMEL, and WikiDiverse demonstrate MECI's superiority over state-of-the-art baselines, while ablation studies validate the efficacy of each proposed component.
\end{itemize}

    
    

\section{Related Work}
\subsection{Text-based Entity Linking}
Text-based Entity Linking (EL) aims to ground ambiguous textual mentions to their corresponding entries in a knowledge base. Generally, existing methodologies fall into two paradigms: local-level and global-level models. Local-level models~\cite{ref2, ref1, ref28, ref43} focus on utilizing the immediate context, such as surrounding words or sentences, to aid disambiguation. In contrast, global-level models~\cite{ref47,ref44,ref45, ref46} incorporate entity or topic coherence to ensure semantic consistency between mentions and entities. Nevertheless, these unimodal methods often struggle to effectively handle multimodal data that combines both textual and visual information.

\begin{figure*}[t] 
    \centering
    \includegraphics[width=\linewidth, height=10cm]{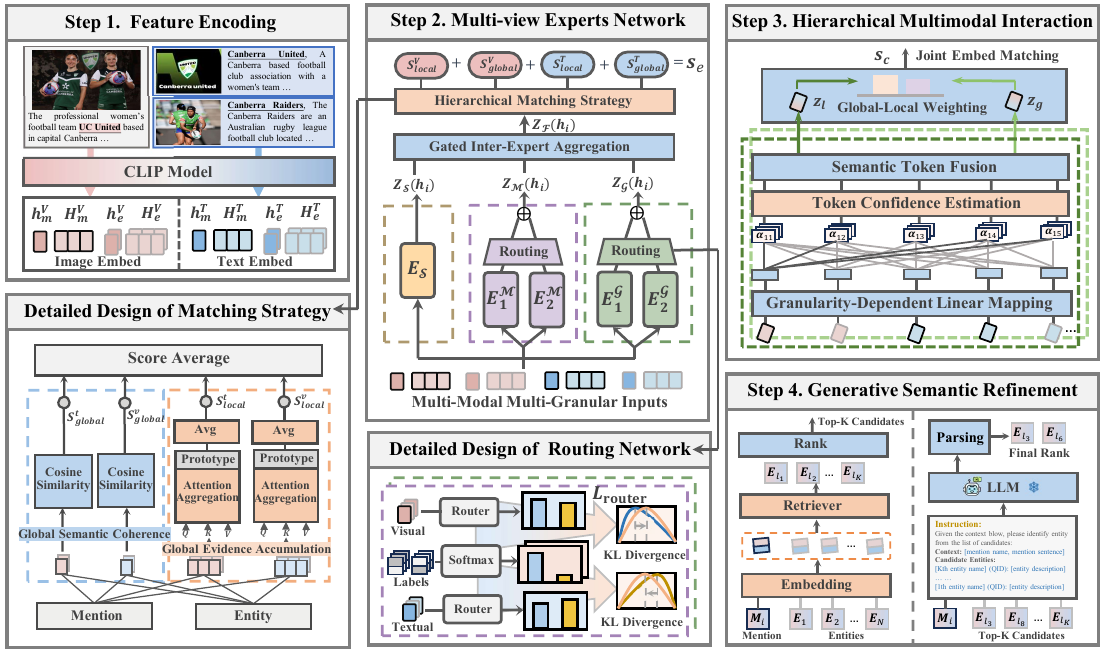} 
    \caption{The Overall Framework of MECI, illustrating a sequential workflow: (1) Feature Encoding $\rightarrow$ (2) Multi-view Experts Network $\rightarrow$ (3) Hierarchical Multimodal Interaction $\rightarrow$ (4) Generative Semantic Refinement.}
    \label{model}
\end{figure*}

\subsection{Multimodal Entity Linking}
Multimodal Entity Linking (MEL) extends traditional entity linking by leveraging visual context for disambiguation. Its main goal is to map mentions in multimodal content to specific entities in a multimodal knowledge base. To address this, existing studies have proposed various sophisticated mechanisms, which can be broadly categorized into distinct paradigms. First, hierarchical gating mechanisms~\cite{ref12,ref13,ref14,ref24} and attention-based paradigms~\cite{ref15,ref17,ref3} aim to refine feature matching through multi-level structures or cross-modal attention. However, these methods often employ coupled interaction designs, limiting their adaptability to varying input semantics and leading to element dominance. Meanwhile, graph-based aggregation schemes~\cite{ref19,ref20,ref21} explicitly model cross-modal relations as graphs, whereas adaptive fusion strategies~\cite{ref23,ref22} aggregate features into unified representations. Despite their utility, these approaches tend to fuse information into global representations, making it difficult to capture fine-grained, context-dependent visual-textual associations. Recently, LLM-based methods~\cite{ref31,ref32,ref4} have achieved remarkable progress by leveraging the generative capabilities of large models, yet remain significantly constrained by excessive computational resource consumption. In summary, existing methods exhibit limitations in comprehensively modeling multi-element collaboration across modalities and granularities. Current designs tend to over-rely on prominent elements, causing modality dominance and granularity bias, which ultimately undermines linking performance in complex multimodal scenarios.

\section{Methodology}

\subsection{Problem Formulation}
MEL aims to resolve a mention $m$ to its target entity $e$ from a knowledge base $\mathcal{E}$. Formally, an entity is defined as a tuple $e \triangleq (e_t, e_v)$, consisting of textual attributes $e_t$ and a visual representation $e_v$. Similarly, a mention is modeled as $m \triangleq (m_t, m_v)$, containing the textual context $m_t$ and an associated image $m_v$. To align mentions with entities, the model parameters $\theta$ are optimized to maximize the likelihood of the ground-truth on the training set $\mathcal{D}$:
\begin{equation}
\theta^* = \arg\max_{\theta} \sum_{(m,e) \in \mathcal{D}} \log p_\theta(e | m, \mathcal{E})
\end{equation}
Here, $p_\theta(e | m, \mathcal{E})$ serves as a similarity metric to evaluate the relevance of a candidate entity $e$ to the query mention $m$.

\subsection{Feature Encoding}
To initialize the representations for the entity $e$ and mention $m$, the model encodes their associated textual and visual information, as illustrated in Figure \ref{model} (step 1). Specifically, the framework leverages the pre-trained CLIP model (specifically \texttt{CLIP-ViT-Base-Patch32}). We utilize its text encoder ($\text{Enc}^T$) for textual feature extraction and its image encoder ($\text{Enc}^V$) for visual feature encoding:
\begin{equation}
\begin{aligned}
\boldsymbol{H}_m^T, \boldsymbol{H}_m^V &= \text{Enc}^T(m_t),\text{Enc}^V(m_v) \\
\boldsymbol{H}_e^T, \boldsymbol{H}_e^V &= \text{Enc}^T(e_t), \text{Enc}^V(e_v)
\end{aligned}
\label{fea_enc}
\end{equation}
where $\boldsymbol{H}_m^T \in \mathbb{R}^{L_1 \times d}$ and $\boldsymbol{H}_e^T \in \mathbb{R}^{L_2 \times d}$ represent the token-level textual sequences (lengths $L_1, L_2$), while $\boldsymbol{H}_m^V \in \mathbb{R}^{P_1 \times d}$ and $\boldsymbol{H}_e^V \in \mathbb{R}^{P_2 \times d}$ denote the patch-level visual sequences (patch counts $P_1, P_2$). Beyond these fine-grained representations, we extract coarse-grained global vectors $\boldsymbol{h}_m^T, \boldsymbol{h}_e^T, \boldsymbol{h}_m^V, \boldsymbol{h}_e^V$ to capture comprehensive semantic context. These are obtained via the output of the special [EOS] token from $\text{Enc}^T$ and the [CLS] token from $\text{Enc}^V$, respectively.

\subsection{Multi-view Experts Network}

To effectively model multi-element correspondences between mentions and entities, we recognize that inputs of varying modalities and granularities offer diverse and complementary evidence from distinct perspectives. However, existing strategies, such as feature-coupled interactions  \cite{ref17, ref24} or global alignment \cite{ref23,ref22}, often suffer from element dominance, thereby struggling to comprehensively capture these dependencies.

To address these challenges, we propose the \textbf{Multi-view Experts Network (MVEN)}. By adopting a decoupled representation learning paradigm, MVEN ensures specialized processing capabilities for heterogeneous features, thereby effectively mitigating element dominance. Specifically, MVEN comprises three key components: (1) a \textbf{View-Specific Experts Module} employing a ``Divide-and-Conquer'' strategy to capture universal patterns and view-specific characteristics; (2) a \textbf{KL-Guided Routing Mechanism} introducing structural priors to balance expert specialization and knowledge sharing; and (3) a \textbf{Gated Aggregation and Matching Layer} that dynamically fuses expert signals and performs specialized correlations computation across diverse perspectives.
\enlargethispage{0.2\baselineskip}
\subsubsection{View-Specific Experts Module}
To capture semantic information, we adopt a ``Divide-and-Conquer'' strategy to model both common patterns and view-specific characteristics.

\textbf{Universal Semantic Projection.}
Formally, we formulate each expert unit $E(\cdot)$ as a learnable transformation comprising a linear projection, normalization, and activation. Let $\boldsymbol{H} = [\boldsymbol{h}^T; \boldsymbol{H}^T; \boldsymbol{h}^V; \boldsymbol{H}^V] \in \mathbb{R}^{N \times d}$ denote the multimodal input features for either a mention or an entity. Based on this, we employ a shared expert mechanism to extract view-agnostic semantic structures. Accordingly, for any arbitrary input token $\boldsymbol{h}_i \in \mathbb{R}^d$, the universal expert $E_{u}(\cdot)$ projects it into a unified semantic space:
\begin{equation}
      \boldsymbol{z}_{\mathcal{S}}(\boldsymbol{h}_i) = E_{u}(\boldsymbol{h}_i) = g\left(\text{LayerNorm}( \boldsymbol{h}_i\boldsymbol{W}_u + \boldsymbol{b}_u)\right)
\end{equation}
where $\boldsymbol{W}_u$ and $\boldsymbol{b}_u$ denote the learnable parameters, and $g(\cdot)$ represents a non-linear activation function (such as ReLU).

\textbf{Collaborative Intra-Expert Fusion.}
To capture the inherent characteristics unique to each perspective, we establish two distinct expert pools: a modality expert set $\Phi^{\mathcal{M}} = \{E^{\mathcal{M}}_m\}_{m=1}^{K_m}$ and a granularity expert set $\Phi^{\mathcal{G}} = \{E^{\mathcal{G}}_g\}_{g=1}^{K_g}$. Despite the greater number of fine-grained tokens, we maintain an equal number of experts for both granularities. This symmetric design ensures balanced interaction and prevents potential fusion bias from imbalanced expert scales. To enrich the representation of a specific view without compromising its distinctiveness, we propose a Collaborative Intra-Expert Fusion mechanism. For a given sample $\boldsymbol{h}_i$ belonging to category $\mathcal{V} \in \{\mathcal{M}, \mathcal{G}\}$, let $v$ denote its ground-truth view index (e.g., a specific modality ID). The fused representation $\boldsymbol{z}_{\mathcal{V}}(\boldsymbol{h}_i)$ is computed as a dynamic weighted aggregation:
\begin{equation}
    \boldsymbol{z}_{\mathcal{V}}(\boldsymbol{h}_i) = \beta_{\mathcal{V}}^{(i)} E^{\mathcal{V}}_v(\boldsymbol{h}_i) + \frac{1-\beta_{\mathcal{V}}^{(i)}}{|\Phi^{\mathcal{V}}|-1} \sum_{E^{\mathcal{V}}_k \in \Phi^{\mathcal{V}}, k \neq v} E^{\mathcal{V}}_k(\boldsymbol{h}_i)
\end{equation}
where $E^{\mathcal{V}}_v$ represents the \textbf{native expert} corresponding to the sample's identity, and the summation term incorporates complementary knowledge from \textbf{peer experts} within the same pool. The coefficient $\beta_{\mathcal{V}}^{(i)} \in (0, 1)$ acts as a dynamic trade-off parameter, governing the balance between preserving native specificity and incorporating external knowledge.

\subsubsection{KL-Guided Routing Mechanism}
The efficacy of the above collaborative fusion depends significantly on the routing mechanism that determines the trade-off coefficient $\beta_{\mathcal{V}}^{(i)}$. Thus, we introduce a KL-Guided Routing Mechanism, a novel approach that leverages structural priors to regularize the routing distribution. Specifically, for each input token $\boldsymbol{h}_i$, we first identify its categorical identity: let $y_{\mathcal{M}}^{(i)}$ and $y_{\mathcal{G}}^{(i)}$ denote its ground-truth modality and granularity indices, respectively. Instead of employing these as rigid one-hot targets, we construct ``Soft Target'' distributions $\mathcal{T}$ to serve as regularizers. This is achieved by applying temperature scaling $\tau$ to the one-hot vectors $\mathbf{e}_y$:
\begin{equation}
    \mathcal{T}_{\mathcal{M}}^{(i)} = \text{Softmax}\left(\frac{\mathbf{e}_{y_{\mathcal{M}}^{(i)}}}{\tau_{\mathcal{M}}}\right), \quad
    \mathcal{T}_{\mathcal{G}}^{(i)} = \text{Softmax}\left(\frac{\mathbf{e}_{y_{\mathcal{G}}^{(i)}}}{\tau_{\mathcal{G}}}\right)
\end{equation}
where $\mathbf{e}_{y}$ denotes the one-hot vector for index $y$. The temperature $\tau$ modulates the prior's focus: a lower $\tau$ sharpens the distribution to enforce \textbf{expert specialization}, while a higher $\tau$ flattens it to encourage \textbf{inter-expert collaboration} and knowledge sharing. 

Simultaneously, for each input $\boldsymbol{h}_i$, the router predicts the routing distributions $\mathcal{P}_{\mathcal{V}}$ via learnable projections:
\begin{equation}
    \mathcal{P}_{\mathcal{V}}^{(i)} = \text{Softmax}\left(\boldsymbol{h}_i \boldsymbol{W}_{\mathcal{V}} + \boldsymbol{b}_{\mathcal{V}}\right), \quad \mathcal{V} \in \{\mathcal{M}, \mathcal{G}\}
\end{equation}
where $\boldsymbol{W}_{\mathcal{V}}$ and $\boldsymbol{b}_{\mathcal{V}}$ are learnable parameters of the routing network.

To align the learned policy with our structural priors while permitting data-driven adaptation, we minimize the Kullback-Leibler (KL) divergence between the predicted distribution $\mathcal{P}$ and the soft target $\mathcal{T}$:
\begin{equation}
\begin{aligned}
\mathcal{L}_{\text{route}} &= \frac{1}{2N} \sum_{i=1}^N \left[ D_{\text{KL}}\left( \mathcal{T}_{\mathcal{M}}^{(i)} \middle\| \mathcal{P}_{\mathcal{M}}^{(i)} \right) + D_{\text{KL}}\left( \mathcal{T}_{\mathcal{G}}^{(i)} \middle\| \mathcal{P}_{\mathcal{G}}^{(i)} \right) \right] \\
&= \frac{1}{2N} \sum_{i=1}^{N} \sum_{\mathcal{V} \in \{\mathcal{M}, \mathcal{G}\}} \sum_{k=1}^{K_\mathcal{V}} \mathcal{T}_{\mathcal{V},k}^{(i)} \log \frac{\mathcal{T}_{\mathcal{V},k}^{(i)}}{\mathcal{P}_{\mathcal{V},k}^{(i)}}
\end{aligned}
\end{equation}
where $K_{\mathcal{V}}$ denotes the number of experts in category $\mathcal{V}$ (i.e., $K_{\mathcal{M}}$ and $K_{\mathcal{G}}$). 

This formulation aligns the router with semantic priors while preserving input context-driven flexibility. By softening $\mathcal{T}$ via the temperature mechanism, the model assigns non-zero probabilities to peer experts, thus avoiding the knowledge isolation typical of hard routing. Finally, the dynamic trade-off coefficient is instantiated as the predicted probability of the native expert: $\beta_{\mathcal{V}}^{(i)} = \mathcal{P}_{\mathcal{V}, v}^{(i)}$, where $v = y_{\mathcal{V}}^{(i)}$.

\subsubsection{Gated Inter-Expert Aggregation}
Building upon the universal ($\boldsymbol{z}_{\mathcal{S}}$), modality-specific ($\boldsymbol{z}_{\mathcal{M}}$), and granularity-specific ($\boldsymbol{z}_{\mathcal{G}}$) representations, we introduce a Gated Inter-Expert Aggregation strategy to integrate these heterogeneous signals. Specifically, for an input token $\boldsymbol{h}_i$, we determine the salience of each expert category $\mathcal{V} \in \{\mathcal{M}, \mathcal{G}\}$ by computing gating coefficients:
\begin{equation}
    \lambda_{\mathcal{V}}^{(i)} = \sigma\left((\boldsymbol{w}_{\mathcal{V}}^{g})^\top \boldsymbol{z}_{\mathcal{V}}(\boldsymbol{h}_i) + b_{\mathcal{V}}^{g}\right)
\end{equation}
These coefficients adaptively modulate expert contributions to produce the fused representation:
\begin{equation}
    \boldsymbol{z}_{\mathcal{F}}(\boldsymbol{h}_i) = \boldsymbol{z}_{\mathcal{S}}(\boldsymbol{h}_i) + \lambda_{\mathcal{M}}^{(i)} \boldsymbol{z}_{\mathcal{M}}(\boldsymbol{h}_i) + \lambda_{\mathcal{G}}^{(i)} \boldsymbol{z}_{\mathcal{G}}(\boldsymbol{h}_i)
\end{equation}
This residual-like formulation maintains a stable universal foundation while flexibly incorporating evidence tailored to the specific context. Consequently, the final representations for a mention $m$ and a candidate entity $e$ are constructed as the concatenation of all view-specific features: $\boldsymbol{Z}_{m/e} = [\boldsymbol{z}^T; \boldsymbol{Z}^T; \boldsymbol{z}^V; \boldsymbol{Z}^V]_{m/e}$.

\subsubsection{Hierarchical Matching Strategy}
To accommodate the distinct matching requirements of heterogeneous perspectives, we categorize the full view set into global views $\Omega_g = \{T_g, V_g\}$ and local views $\Omega_l = \{T_l, V_l\}$, where $T$ and $V$ denote textual and visual modalities, and subscripts $g$ and $l$ represent global and local granularities, respectively. For global views $v \in \Omega_g$, we prioritize \textbf{Global Semantic Coherence}. Given the global feature vectors $\boldsymbol{z}_m^v$ and $\boldsymbol{z}_e^v$, we compute their correlation via cosine similarity:
\begin{equation}
    s_{\mathrm{global}}^v = \cos(\boldsymbol{z}_m^v, \boldsymbol{z}_e^v) = \frac{(\boldsymbol{z}_m^v)^\top \boldsymbol{z}_e^v}{\|\boldsymbol{z}_m^v\|_2\,\|\boldsymbol{z}_e^v\|_2}
\end{equation}
Conversely, for fine-grained views $v \in \Omega_l$, we employ a \textbf{Local Evidence Accumulation} layer. Given local feature sequences $\boldsymbol{Z}_m^v$ and $\boldsymbol{Z}_e^v$, we first construct context-aware entity prototypes $\tilde{\boldsymbol{Z}}_e^v$ via attention-based aggregation: $\tilde{\boldsymbol{Z}}_e^v = \text{Softmax}(\boldsymbol{Z}_m^v (\boldsymbol{Z}_e^v)^\top / \sqrt{d}) \boldsymbol{Z}_e^v$. This mechanism dynamically aligns entity features with the mention context, effectively highlighting relevant evidence segments while suppressing noise. The local score is then obtained by averaging the cosine similarity at each local position:
\begin{equation}
    s_{\mathrm{local}}^v = \frac{1}{L_v} \sum_{i=1}^{L_v} \frac{(\boldsymbol{z}_{m,i}^v)^\top \tilde{\boldsymbol{z}}_{e,i}^v}{\|\boldsymbol{z}_{m,i}^v\|_2\,\|\tilde{\boldsymbol{z}}_{e,i}^v\|_2}
\end{equation}
Finally, the comprehensive similarity $s_e$ is derived by averaging the matching scores across all global and local views:
\begin{equation}
    s_e = \frac{\displaystyle\sum_{v\in\Omega_g}s_{\mathrm{global}}^v + \displaystyle\sum_{v\in\Omega_l}s_{\mathrm{local}}^v}{|\Omega_g|+|\Omega_l|}
\end{equation}

\subsection{Hierarchical Multimodal Interaction}
While the Multi-view Experts Network effectively models mention-entity correlations through decoupled representations, these view-specific features remain relatively isolated, limiting the model's ability to capture multi-element correspondences between modalities. To bridge this gap, we develop the Hierarchical Multimodal Interaction Module (HMIM). HMIM first employs a \textbf{Dynamic Cross-modal Weighting Network (DCWN)} to refine interactions at distinct granularities. Subsequently, a \textbf{Granularity-aware Gating Network} dynamically fuses the multi-granular evidence into a robust joint representation for precise matching.

\subsubsection{Dynamic Cross-modal Weighting Network}
The DCWN processes fine-grained and coarse-grained streams in parallel to model intricate cross-modal dependencies. This hierarchical strategy effectively avoids the granularity bias inherent in unified interaction paradigms, where multi-level features are concatenated without structural distinction. We describe the operation using the fine-grained token stream as an illustrative example. Let $\boldsymbol{H} = [\boldsymbol{Z}^T; \boldsymbol{Z}^V] \in \mathbb{R}^{L \times d}$ denote the input sequence combining textual and visual local tokens derived from the previous module. We employ a multi-head cross-modal attention mechanism with $N_h$ heads to model pairwise correspondences. For the $i$-th head ($i = 1, \dots, N_h$), the input is projected into query, key, and value subspaces via learnable matrices: $\boldsymbol{Q}^{(i)} = \boldsymbol{H} \boldsymbol{W}_Q^{(i)}$, $\boldsymbol{K}^{(i)} = \boldsymbol{H} \boldsymbol{W}_K^{(i)}$, and $\boldsymbol{V}^{(i)} = \boldsymbol{H} \boldsymbol{W}_V^{(i)}$, where $\boldsymbol{W}_Q^{(i)}, \boldsymbol{W}_K^{(i)}, \boldsymbol{W}_V^{(i)} \in \mathbb{R}^{d \times d_h}$.

The attention matrix $\boldsymbol{A}^{(i)} \in \mathbb{R}^{L \times L}$ captures the pairwise correspondences between tokens and is computed as:
\begin{equation}
    \boldsymbol{A}_{n,j}^{(i)} = \frac{
        \exp\left( (\boldsymbol{Q}_n^{(i)})^\top \boldsymbol{K}_j^{(i)} / \sqrt{d_h} \right)
    }{
        \sum_{k=1}^{L} \exp\left( (\boldsymbol{Q}_n^{(i)})^\top \boldsymbol{K}_k^{(i)} / \sqrt{d_h} \right)
    }
\end{equation}
The context-enriched output is derived by aggregating the values weighted by $\boldsymbol{A}^{(i)}$, followed by concatenation and a linear projection $\boldsymbol{W}_O$:
\begin{equation}
    \tilde{\boldsymbol{H}} = \text{LayerNorm}\left( \left( \bigoplus_{i=1}^{N_h} \boldsymbol{A}^{(i)} \boldsymbol{V}^{(i)} \right) \boldsymbol{W}_O + \boldsymbol{H} \right)
\end{equation}
Crucially, to explicitly quantify the reliability of cross-modal correspondences and refine the features, we introduce a \textbf{Token Confidence Estimation (TCE)} mechanism. Rather than treating all tokens equally, TCE computes a scalar confidence score $w_n$ for each token $n$, reflecting its saliency within the multimodal context. This is formulated by aggregating the attention scores across all heads and tokens:
\begin{equation}
    w_n = \frac{
        \exp\left( \frac{1}{\sqrt{L \cdot N_h}} \sum_{j=1}^{L} \sum_{i=1}^{N_h} \boldsymbol{A}_{j,n}^{(i)} \right)
    }{
        \sum_{k=1}^{L} \exp\left( \frac{1}{\sqrt{L \cdot N_h}} \sum_{j=1}^{L} \sum_{i=1}^{N_h} \boldsymbol{A}_{j,k}^{(i)} \right)
    }
\end{equation}
This mechanism acts as a soft attention filter, adaptively emphasizing tokens that exhibit strong semantic consistency. The final granularity-specific representation is obtained by re-weighting the enhanced features:
\begin{equation}
    \boldsymbol{z}_{l} = \sum_{n=1}^{L} w_n^l \tilde{\boldsymbol{h}}_n^l, \quad
    \boldsymbol{z}_{g} = \sum_{n=1}^{L'} w_n^g \tilde{\boldsymbol{h}}_n^g
\end{equation}
where $\boldsymbol{z}_{l}$ and $\boldsymbol{z}_{g}$ denote the aggregated fine-grained (local) and coarse-grained (global) representations, respectively (with $L'$ denoting the length of the global stream).

\subsubsection{Granularity-aware Gating Network}
To effectively integrate the refined multi-granular evidence from $\boldsymbol{z}_{g}$ and $\boldsymbol{z}_{l}$, the Granularity-aware Gating Network dynamically fuses them into a robust joint representation. We compute granularity-aware gating scores $\alpha_k$ for $k \in \{g, l\}$ via:
\begin{equation}
    \alpha_k = \frac{
        \exp\left( \boldsymbol{w}_g^\top \tanh(\boldsymbol{W}_k \boldsymbol{z}_{k}) \right)
    }{
        \sum_{j \in \{g,l\}} \exp\left( \boldsymbol{w}_g^\top \tanh(\boldsymbol{W}_j \boldsymbol{z}_{j}) \right)
    }
\end{equation}
where $\boldsymbol{W}_k$ and $\boldsymbol{w}_g$ are learnable parameters. The unified representation $\boldsymbol{u}$ is then derived through a weighted aggregation:
\begin{equation}
    \boldsymbol{u} = \alpha_g \boldsymbol{z}_{g} + \alpha_l \boldsymbol{z}_{l}
\end{equation}
This mechanism enables the model to selectively emphasize the granularity that is most informative for the current context. Finally, the matching score is computed as the cosine similarity between the unified mention and entity representations:
\begin{equation}
    s_{c} = \cos(\boldsymbol{u}_{m}, \boldsymbol{u}_{e}) = \frac{\boldsymbol{u}_{m}^\top \boldsymbol{u}_{e}}{\|\boldsymbol{u}_{m}\| \|\boldsymbol{u}_{e}\|}
\end{equation}

\subsection{Optimization and Generative Refinement}
\subsubsection{Holistic Training Objectives}
We employ a multi-perspective contrastive learning paradigm \cite{ref14} to enforce discriminative alignment across both individual and fused perspectives. Specifically, we apply contrastive constraints to three distinct scoring components: the expert-based score $s_e$, the interaction-based score $s_c$, and the overall consensus score $s_o = s_e + s_c$. The InfoNCE loss $\mathcal{L}_{\phi}$ for a given scoring function $s_\phi$ (where $\phi \in \{e, c, o\}$) is defined as:
\begin{equation}
    \mathcal{L}_{\phi} = -\mathbb{E} \left[ \log \frac{
        \exp(s_\phi(m, e))
    }{
        \exp(s_\phi(m, e)) + \sum_{e' \in \mathcal{N}_m} \exp(s_\phi(m, e'))
    } \right]
\end{equation}
where $\mathcal{N}_{m}$ denotes the set of negative samples corresponding to the anchor $m$. The total objective aggregates these matching losses with the routing regularization term:
\begin{equation}
    \mathcal{L}_{\text{total}} = \sum_{\phi \in \{e, c, o\}} \mathcal{L}_{\phi} + \lambda \, \mathcal{L}_{\text{route}}
\end{equation}
where $\lambda$ governs the weight of the KL-divergence constraint. This formulation ensures that the model learns robust representations while strictly adhering to the semantic structural priors imposed on the router.

\subsubsection{LLM-based Generative Reasoning}
Inspired by~\cite{ref4, ref29}, we incorporate a \textbf{Generative Semantic Refinement} stage to further enhance entity linking precision. Initially, a candidate set $\mathcal{C}_m$ is retrieved based on the top-$K$ scores $s_o$:
\begin{equation}
    \mathcal{C}_m = \operatorname*{arg\,topK}_{e \in \mathcal{E}} \big( s_o(m, e) \big)
\end{equation}
Subsequently, the refinement stage is reformulated as a multiple-choice problem. By constructing a prompt $\mathcal{I}(m, \mathcal{C}_m)$ that integrates the mention context $t_m$ and candidate descriptions $\{t_e\}$—as illustrated in the prompt design in Figure~\ref{model} (step 4)—the optimal entity $e^*$ is determined by maximizing the LLM's generation probability:
\begin{equation}
    e^* = \operatorname*{argmax}_{e \in \mathcal{C}_m} \, \mathbb{P}_{\text{LLM}}\Big( e \mid \mathcal{I}(m, \mathcal{C}_m); \Theta \Big)
\end{equation}
where $\Theta$ denotes frozen LLM parameters. This way effectively leverages parametric knowledge to pinpoint the ground-truth entity.

\section{Evaluation}
In this section, we evaluate our proposed MECI through comprehensive experiments on three public MEL datasets, accompanied by extensive analyses to provide deeper insights into our framework.

\subsection{Experimental Setup}
\subsubsection{Datasets}
To evaluate our approach, we utilize three established datasets: WikiDiverse~\cite{ref34}, WikiMEL~\cite{ref12}, and RichpediaMEL~\cite{ref12}. Derived from Wikipedia entity pages, WikiMEL targets Wikidata as its knowledge base and consists of over 22,000 multimodal sentences, focusing predominantly on Person entities. In contrast, WikiDiverse is sourced from Wikinews and maps to Wikipedia; it spans seven distinct entity categories, including Person, Organization, Location, Country, Event, Works, and Misc. RichpediaMEL integrates Wikidata attributes for entities found in Richpedia~\cite{ref35} with multimodal content from Wikipedia. Regarding data partitioning, we adopt the standard split configurations: an 8:1:1 ratio for WikiDiverse, and a 7:1:2 ratio for both WikiMEL and RichMEL. The statistics of these datasets are shown in Table \ref{tab1}.

\subsubsection{Baselines and Evaluation Metrics}
We evaluate our method against three groups of representative baselines: (1) Text-based EL methods, such as BERT~\cite{ref1}, RoBERTa~\cite{ref25-1}, and BLINK~\cite{ref2}; (2) Discriminative MEL methods, covering CLIP~\cite{ref36}, ViLT~\cite{ref37}, ALBEF~\cite{ref38}, METER~\cite{ref39}, GHMFC~\cite{ref12}, MIMIC~\cite{ref14}, OT-MEL~\cite{ref40}, MELOV~\cite{ref13}, M\textsuperscript{3}EL~\cite{ref3}, and MMOE~\cite{ref24}; and (3) LLM-based methods, including GEMEL~\cite{ref31}, UniMEL~\cite{ref29}, IIER~\cite{ref17}, and I2CR~\cite{ref4}. To assess model performance, we utilize two standard metrics: Mean Reciprocal Rank (MRR) and Hits@$N$. MRR calculates the average of the reciprocal ranks of the first correct answer. Hits@$N$ reports the percentage of test cases where the correct entity appears within the top $N$ predictions, where $N \in \{1, 3, 5\}$.

\subsubsection{Implementations}
We conducted the experiments on a workstation equipped with an Intel Core i9-12900K CPU (64GB RAM) and two NVIDIA A100 GPUs (2x40GB). we employ the pre-trained CLIP-Vit-Base-Patch32~\cite{ref36} model as the initialization of the textual and visual encoder.  For input preprocessing, visual inputs were resized to a resolution of $224 \times 224$. Subsequently, the maximum sequence length was set to 50 tokens for both visual and textual inputs. A unified training configuration was maintained across all datasets. The models were optimized using AdamW \cite{ref41} with a batch size of 64 for 20 epochs. The learning rate was initialized at $5 \times 10^{-6}$ for WikiMEL and $1\times10^{-5}$ for both the WikiDiverse and RichpediaMEL datasets. In the re-ranking phase, we utilized DeepSeek-v3 to re-rank the retrieved candidates, setting the candidate pool size to $K=16$. Our code is available at \href{https://github.com/pengjiebig/MEL-MECI-main}{MEL-MECI-main}.

\begin{table}[t]
\centering
\caption{Statistics of three MEL datasets.}
\label{tab1}
\renewcommand{\arraystretch}{0.92}
\resizebox{\linewidth}{!}{
\begin{tabular}{l|ccc}
\toprule
\textbf{Statistics} & \textbf{WikiDiverse} & \textbf{WikiMEL} & \textbf{RichpediaMEL} \\
\midrule
\# number of sentences & 7,405 & 22,070 & 17,724 \\
\# number of entities & 132,460 & 109,976 & 160,935 \\
\# entities with image & 67,309 & 67,195 & 86,769 \\
\# number of mentions & 15,093 & 25,846 & 17,805 \\
\# image of mentions & 6,697 & 22,136 & 15,853 \\
\midrule
\# mentions in train & 11,351 & 18,092 & 12,463 \\
\# mentions in valid & 1,664 & 2,585 & 1,780 \\
\# mentions in test & 2,078 & 5,169 & 3,562 \\
\bottomrule
\end{tabular}
}
\end{table}

\begin{table*}[htbp]
\definecolor{bestbg}{HTML}{E2F0D9}
\caption{Experimental results (\%) on three MEL benchmark datasets: \textbf{WikiDiverse}, \textbf{RichpediaMEL}, and \textbf{WikiMEL}. The \colorbox{bestbg}{\textbf{bold}} scores indicate the best results, and the \underline{underlined} scores represent the second-best results.}
\label{tab2}
\renewcommand{\arraystretch}{0.9}
\centering
\resizebox{\textwidth}{!}{
\begin{tabular}{lc|cccc|cccc|cccc}
\toprule
\multirow{2}{*}{\textbf{Methods}} & \multirow{2}{*}{\textbf{Venue}} & \multicolumn{4}{c}{\textbf{WikiDiverse}} & \multicolumn{4}{c}{\textbf{RichpediaMEL}} & \multicolumn{4}{c}{\textbf{WikiMEL}} \\
\cmidrule(lr){3-6} \cmidrule(lr){7-10} \cmidrule(lr){11-14}
 & & \textbf{MRR} & \textbf{Hits@1} & \textbf{Hits@3} & \textbf{Hits@5} & \textbf{MRR} & \textbf{Hits@1} & \textbf{Hits@3} & \textbf{Hits@5} & \textbf{MRR} & \textbf{Hits@1} & \textbf{Hits@3} & \textbf{Hits@5} \\
\midrule
BERT & NAACL'19 & 67.38 & 55.77 & 75.73 & 83.11 & 71.67 & 59.55 & 81.12 & 87.16 & 81.78 & 74.82 & 86.79 & 90.47 \\
RoBERTa & arXiv'19 & 70.52 & 59.46 & 78.54 & 85.08 & 72.80 & 61.34 & 81.56 & 87.15 & 80.86 & 73.75 & 85.85 & 89.80 \\
BLINK & EMNLP'20 & 69.15 & 57.14 & 78.04 & 85.32 & 71.39 & 58.47 & 81.51 & 88.09 & 81.72 & 74.66 & 86.63 & 90.57 \\
\midrule
CLIP & ICML'21 & 71.69 & 61.21 & 79.63 & 85.18 & 77.57 & 67.78 & 85.22 & 90.04 & 88.23 & 83.23 & 92.10 & 94.51 \\
ViLT & ICML'21 & 45.22 & 34.39 & 51.07 & 57.83 & 56.63 & 45.85 & 62.96 & 69.80 & 79.46 & 72.64 & 84.51 & 87.86 \\
ALBEF & NeurIPS'21 & 69.93 & 60.59 & 75.59 & 81.30 & 75.29 & 65.17 & 82.84 & 88.28 & 84.56 & 78.64 & 88.93 & 91.75 \\
METER & CVPR'22 & 63.71 & 53.14 & 70.93 & 77.59 & 74.15 & 63.96 & 82.24 & 87.08 & 79.49 & 72.46 & 84.41 & 88.17 \\
GHMFC & SIGIR'22 & 70.99 & 60.27 & 79.40 & 84.74 & 80.76 & 72.92 & 86.85 & 90.60 & 83.36 & 76.55 & 88.40 & 92.01 \\
MIMIC & KDD'23 & 73.44 & 63.51 & 81.04 & 86.43 & 86.95 & 81.02 & 91.77 & 94.38 & 91.82 & 87.98 & 95.07 & 96.37 \\
OT-MEL & ACL'24 & 75.43 & 66.07 & 82.82 & 87.39 & 88.27 & 83.30 & 92.39 & 94.83 & 92.59 & 88.97 & 95.63 & 96.96 \\
MELOV & ACL'24 & 76.57 & 67.32 & 83.69 & 87.54 & 88.80 & 84.14 & 92.81 & 94.89 & 92.32 & 88.91 & 95.61 & 96.58 \\
M\textsuperscript{3}EL & KDD'25 & 81.29 & 74.06 & 86.57 & 90.04 & 88.26 & 82.82 & 92.73 & 95.34 & 92.30 & 88.84 & 95.20 & 96.71 \\
MMOE & KDD'25 & 81.72 & 74.59 & 87.25 & 90.33 & 89.04 & 84.36 & 92.93 & 94.64 & 92.53 & 89.07 & 95.36 & 97.02 \\
\midrule
GEMEL & COLING'24 & - & 86.30 & - & - & - & - & - & - & - & 82.60 & - & - \\
UniMEL & CIKM'24 & - & 86.50 & 89.80 & 90.40 & - & 85.20 & 92.60 & 96.80 & - & 88.70 & 95.20 & 96.90 \\
IIER & ACL'25 & - & 69.47 & 84.43 & 88.79 & - & 84.63 & 93.27 & 95.30 & - & 88.93 & 95.69 & 96.73 \\
I2CR & MM'25 & - & \underline{91.60} & \underline{94.70} & \underline{95.60} & - & \underline{86.80} & 92.90 & \underline{97.20} & - & \underline{92.20} & \underline{96.10} & \underline{97.50} \\
\midrule
MECI & - & \underline{86.71} & 79.93 & 90.23 & 92.93 & \underline{89.94} & 85.63 & \underline{93.43} & 95.65 & \underline{93.48} & 90.38 & 96.05 & 97.21 \\
MECI+ & - & \cellcolor{bestbg}\textbf{95.00} & \cellcolor{bestbg}\textbf{93.55} & \cellcolor{bestbg}\textbf{96.01} & \cellcolor{bestbg}\textbf{96.97} & \cellcolor{bestbg}\textbf{95.49} & \cellcolor{bestbg}\textbf{94.10} & \cellcolor{bestbg}\textbf{96.38} & \cellcolor{bestbg}\textbf{97.39} & \cellcolor{bestbg}\textbf{96.25} & \cellcolor{bestbg}\textbf{94.51} & \cellcolor{bestbg}\textbf{97.72} & \cellcolor{bestbg}\textbf{98.34} \\
\bottomrule
\end{tabular}
}
\end{table*}

\subsection{Main Results}
The comparative results on three MEL datasets are summarized in Table \ref{tab2}. Overall, our proposed MECI framework demonstrates superior performance across all benchmarks. Specifically, our MECI consistently outperforms previous strong discriminative models, while our zero-shot generative refinement model MECI+ achieves new state-of-the-art results. The detailed analysis is as follows:

\textbf{$\triangleright$ Comparison with Text-based Methods.}
Text-based EL consistently underperforms multimodal approaches. The best text-only model achieves only 74.82\% Hits@1 on WikiMEL compared to over 90\% for our method. This suggests these text-based EL baselines struggle to capture deep semantics. Our approach addresses this by learning richer representations from both textual context and auxiliary visual signals.

\textbf{$\triangleright$ Comparison with Discriminative MEL Methods.}
Compared to state-of-the-art discriminative baselines like MMOE and M\textsuperscript{3}EL, our \textbf{MECI} achieves consistent superiority, surpassing the strongest baseline MMOE by \textbf{5.34\%, 1.27\%, and 1.31\%} in Hits@1 on WikiDiverse, RichpediaMEL, and WikiMEL, respectively. This comes from our collaborative design: MVEN employs a “divide-and-conquer” strategy to decouple subspaces and mitigate element dominance, while HMIM utilizes a granularity-dependent cross-modal weighting network to capture intricate cross-modal dependencies. In contrast to baselines limited by coupled interactions or coarse global fusion, MECI effectively alleviates element dominance, capturing comprehensive multi-element correspondences.

\textbf{$\triangleright$ Comparison with Generative Methods.}
Finally, we benchmark \textbf{MECI+} (MECI augmented with a zero-shot generative refinement stage) against LLM-based paradigms. While I2CR exhibits competitive performance, our zero-shot MECI+ surpasses the strongest baseline by large margins of \textbf{1.95\%}, \textbf{7.30\%}, and \textbf{2.31\%} in Hits@1 across the three datasets, respectively. Crucially, our model realizes these gains without the need for resource-intensive fine-tuning. By leveraging a frozen DeepSeek V3 to re-rank candidates via a multiple-choice prompting strategy, we demonstrate that a robust retrieval foundation is sufficient to effectively elicit the reasoning capabilities of LLMs. 

\begin{figure}[t]
    \centering
    \includegraphics[width=\linewidth]{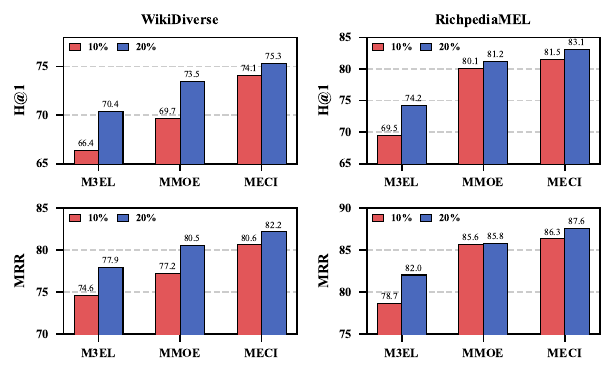}
    \caption{Performance comparison of low resource settings on RichpediaMEL and WikiDiverse datasets.}
    \label{fig3}
\end{figure}

\textbf{$\triangleright$ Low Resource Setting.}
We further examine model performance under data-constrained settings (10\% and 20\% training data) to address the challenge of costly data annotation. As shown in Figure \ref{fig3}, our method consistently surpasses baselines M\textsuperscript{3}EL and MMOE. Note that the MMOE baseline was reproduced by omitting the DME mention text augmentation to ensure comparable input configurations.  Even in the challenging 10\% setting, our framework maintains high performance: it achieves \textbf{74.1\%} accuracy on WikiDiverse (vs. 69.7\% for MMOE) and \textbf{81.5\%} Hits@1 on RichpediaMEL (vs. 80.1\% for MMOE and 69.5\% for M\textsuperscript{3}EL). These significant margins attest to the data efficiency of our approach, highlighting its potential for applications where labeled data is scarce.

\subsection{Ablation Study}
We examine the contribution of specific components through ablation studies on all datasets (Table 3), covering three main aspects:

\begin{table*}[htbp]
\definecolor{bestbg}{HTML}{E2F0D9}
\caption{Ablation studies on three MEL datasets. ``w/o'' denotes the removal of a specific component, and ``repl.'' stands for replacing the component with a variant. The \colorbox{bestbg}{\textbf{bold}} scores indicate the results of the complete model.}
\renewcommand{\arraystretch}{0.9}
\label{tabGOODLUCK}
\centering
\resizebox{\textwidth}{!}{
\begin{tabular}{l|cccc|cccc|cccc}
\toprule
\multirow{2}{*}{\textbf{Methods}} & \multicolumn{4}{c}{\textbf{WikiDiverse}} & \multicolumn{4}{c}{\textbf{RichpediaMEL}} & \multicolumn{4}{c}{\textbf{WikiMEL}} \\
\cmidrule(lr){2-5} \cmidrule(lr){6-9} \cmidrule(lr){10-13}
 & \textbf{MRR} & \textbf{Hits@1} & \textbf{Hits@3} & \textbf{Hits@5} & \textbf{MRR} & \textbf{Hits@1} & \textbf{Hits@3} & \textbf{Hits@5} & \textbf{MRR} & \textbf{Hits@1} & \textbf{Hits@3} & \textbf{Hits@5} \\
\midrule
MECI & \cellcolor{bestbg}\textbf{86.71} & \cellcolor{bestbg}\textbf{79.93} & \cellcolor{bestbg}\textbf{90.23} & \cellcolor{bestbg}\textbf{92.93} & \cellcolor{bestbg}\textbf{89.94} & \cellcolor{bestbg}\textbf{85.63} & \cellcolor{bestbg}\textbf{93.43} & \cellcolor{bestbg}\textbf{95.65} & \cellcolor{bestbg}\textbf{93.48} & \cellcolor{bestbg}\textbf{90.38} & \cellcolor{bestbg}\textbf{96.05} & \cellcolor{bestbg}\textbf{97.21} \\
\midrule
w/o multi-view expert network & 81.34 & 72.47 & 86.28 & 90.52 & 85.16 & 78.95 & 90.15 & 92.62 & 90.40 & 85.97 & 93.98 & 96.09 \\
w/o shared expert & 85.89 & 78.77 & 90.08 & 92.10 & 89.55 & 85.49 & 92.64 & 94.55 & 93.07 & 89.70 & 95.83 & 96.95 \\
w/o modal expert & 84.95 & 77.24 & 89.12 & 91.72 & 89.01 & 84.42 & 92.76 & 94.58 & 92.26 & 88.58 & 95.25 & 96.88 \\
w/o grain expert & 83.27 & 76.61 & 88.59 & 91.48 & 89.07 & 84.76 & 92.61 & 94.27 & 92.56 & 89.06 & 95.43 & 97.09 \\
\midrule
w/o hierarchical multimodal interaction & 85.76 & 78.58 & 89.93 & 92.78 & 89.32 & 85.13 & 92.86 & 94.50 & 91.94 & 88.52 & 94.49 & 96.26 \\
repl. granularity unified interaction & 85.18 & 77.57 & 89.57 & 91.73 & 88.41 & 83.47 & 92.44 & 94.30 & 91.75 & 88.23 & 94.40 & 96.15 \\
\midrule
w/o expert-based matching $L_e$ & 83.28 & 75.17 & 87.73 & 91.00 & 82.13 & 75.16 & 87.34 & 90.79 & 89.00 & 84.13 & 92.78 & 95.09 \\
w/o interaction-based matching $L_c$ & 85.76 & 78.58 & 89.93 & 92.78 & 89.32 & 85.13 & 92.86 & 94.50 & 91.94 & 88.52 & 94.49 & 96.26 \\
w/o overall consensus matching $L_o$ & 83.99 & 75.84 & 88.26 & 91.53 & 89.03 & 84.12 & 93.01 & 95.19 & 92.03 & 88.23 & 95.16 & 96.73 \\
w/o router loss $L_\text{route}$ & 83.39 & 75.17 & 87.92 & 91.38 & 89.40 & 85.02 & 92.90 & 94.75 & 92.68 & 89.27 & 95.51 & 96.92 \\
repl. hard router & 83.43 & 76.75 & 88.48 & 91.92 & 88.78 & 84.22 & 92.63 & 94.55& 93.03 & 89.62 & 95.95 & 97.06 \\
\midrule
w/o visual modality & 85.02 & 77.33 & 89.60 & 91.92 & 89.05 & 84.96 & 92.22 & 93.87 & 89.72 & 85.11 & 93.32 & 95.53 \\
\bottomrule
\end{tabular}}
\end{table*}

\begin{figure}[t]
    \centering
    \includegraphics[width=\linewidth]{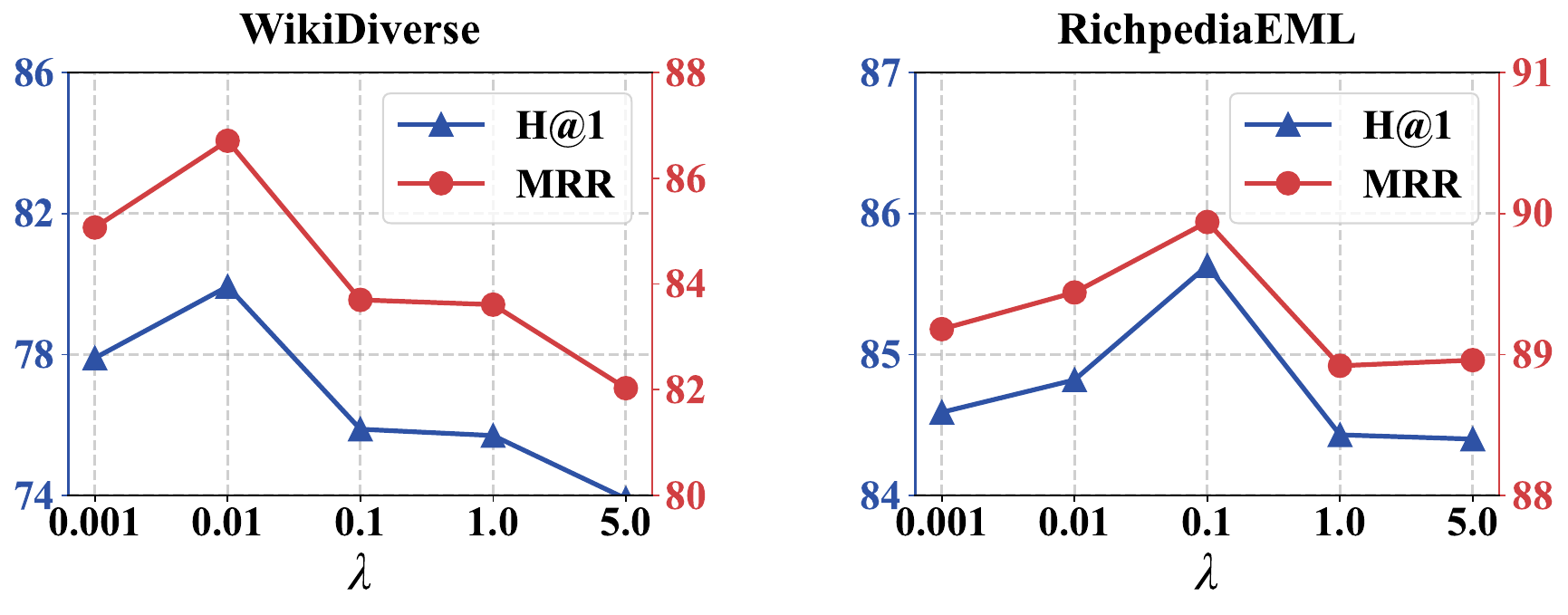}
    \caption{Sensitivity analysis of the routing loss weight $\lambda$.}
    \label{fig4}
\end{figure}

\textbf{$\triangleright$ Impact of Multi-view Experts Network.}
As shown in Table \ref{tabGOODLUCK}, removing the MVEN module leads to the most significant performance degradation, with a sharp drop of 7.46\% in Hits@1 on the WikiDiverse dataset. This empirical evidence strongly supports our ``Divide-and-Conquer'' strategy, indicating that aggregating diverse views is critical for capturing comprehensive evidence. Specifically, ablating the \texttt{grain expert}, which is designed to capture distinctive granularity-specific semantics, causes a Hits@1 decline of 3.32\%. Similarly, the \texttt{modal expert}, responsible for extracting distinctive modality-specific semantics, contributes a notable gain of 2.69\%. Even the \texttt{shared expert}, aimed at modeling common underlying patterns across views to bridge semantic gaps, brings a consistent improvement of 1.16\%. These results confirm that each expert plays an indispensable and complementary role in robust entity linking.

\textbf{$\triangleright$ Effectiveness of Hierarchical Interaction.} Removing the HMIM module (\texttt{w/o hierarchical multimodal interaction}) results in an observable performance drop, e.g., decreasing Hits@1 by 1.86\% on WikiMEL. This indicates that the HMIM module is essential for modeling deep cross-modal dependencies. Crucially, replacing our hierarchical design with a flat interaction mechanism (\texttt{repl. granularity unified interaction})—where fine-grained and coarse-grained features are concatenated for unified DCWN attention processing—yields suboptimal results with a 2.15\% deficit on WikiMEL. This validates the effectiveness of our granularity-specific interaction module, avoiding the granularity bias caused by unified processing.

\textbf{$\triangleright$ Analysis of Optimization and Visual Modality.} Regarding optimization, for instance, the sharp 10.47\% Hits@1 drop on RichpediaMEL without $\mathcal{L}_e$ (\texttt{w/o expert-based matching}) emphasizes the need for direct supervision on expert-based matching, while declines without $\mathcal{L}_c$ and $\mathcal{L}_o$ validate our interaction-based matching and overall consensus matching objective. Notably, removing $\mathcal{L}_c$ and removing the HMIM module correspond to the same ablation setting. Crucially, omitting $\mathcal{L}_{\text{route}}$ (\texttt{w/o router loss}) confirms that KL-guided priors are essential for regularization. These priors enforce a critical balance between specialization and collaboration that unconstrained adaptation lacks. Furthermore, the performance drop with a hard router (\texttt{repl. hard router}) validates our approach: its discrete nature hinders effective expert combination, whereas our differentiable gating smoothly synthesizes diverse capabilities. Finally, the consistent performance drop without visual features (\texttt{w/o visual modality})—exemplified by a 5.27\% decrease on WikiMEL—highlights the necessity of complementary visual evidence.

\subsection{Parameter Sensitivity Analysis}
\hspace*{\widthof{ab}}\textbf{$\triangleright$ Effect of Loss Weight $\lambda$.}
Figure \ref{fig4} presents the impact of the routing loss weight $\lambda$ on model performance. We observe that the performance improves as $\lambda$ increases from small values, confirming the effectiveness of our KL-guided routing mechanism in regulating expert behaviors. The model achieves optimal results at $\lambda=0.01$ for WikiDiverse and $\lambda=0.1$ for RichpediaEML. However, further increasing $\lambda$ leads to performance degradation. This is likely because an excessive weight on $\mathcal{L}_{\text{route}}$ over-constrains the routing network, hindering the model's ability to learn flexible representations for entity matching.

\begin{figure}[t]
    \centering
    \includegraphics[width=\linewidth]{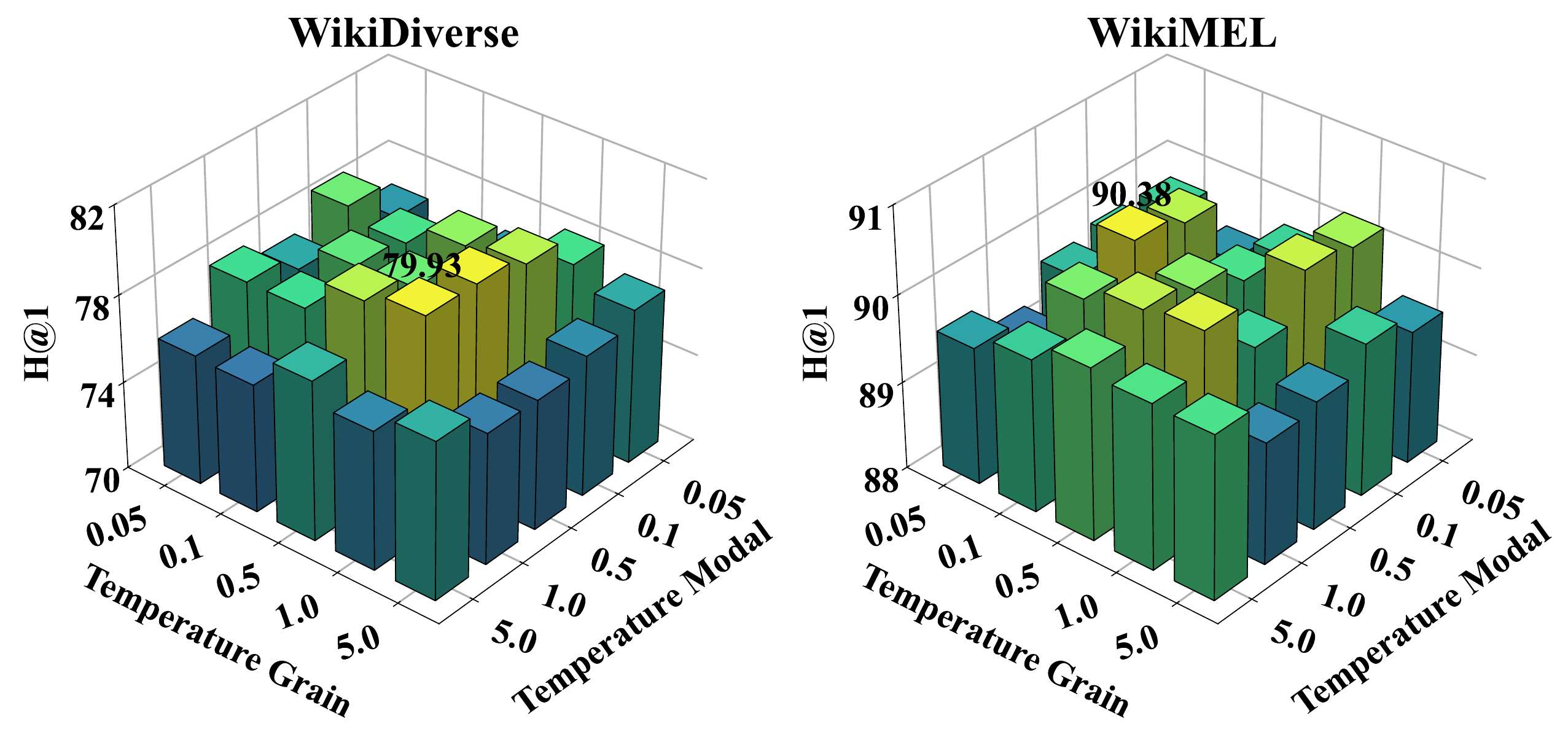}
    \caption{Sensitivity analysis of the temperature $\tau_{\mathcal{M}}$ and $\tau_{\mathcal{G}}$ on the WikiDiverse and WikiMEL datasets.}
    \label{fig5}
\end{figure}

\begin{figure}[t]
    \centering
    \includegraphics[width=\linewidth]{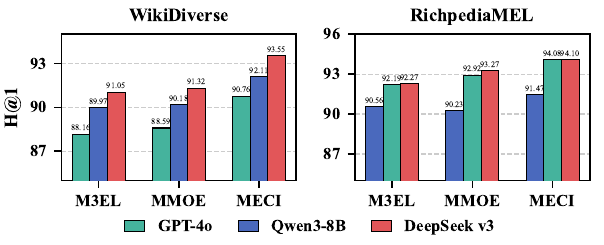}
    \caption{Results of generative semantic refinement stage with different LLMs.}
    \label{fig6}
\end{figure}

\begin{figure}[t]
    \centering
    \includegraphics[width=\linewidth]{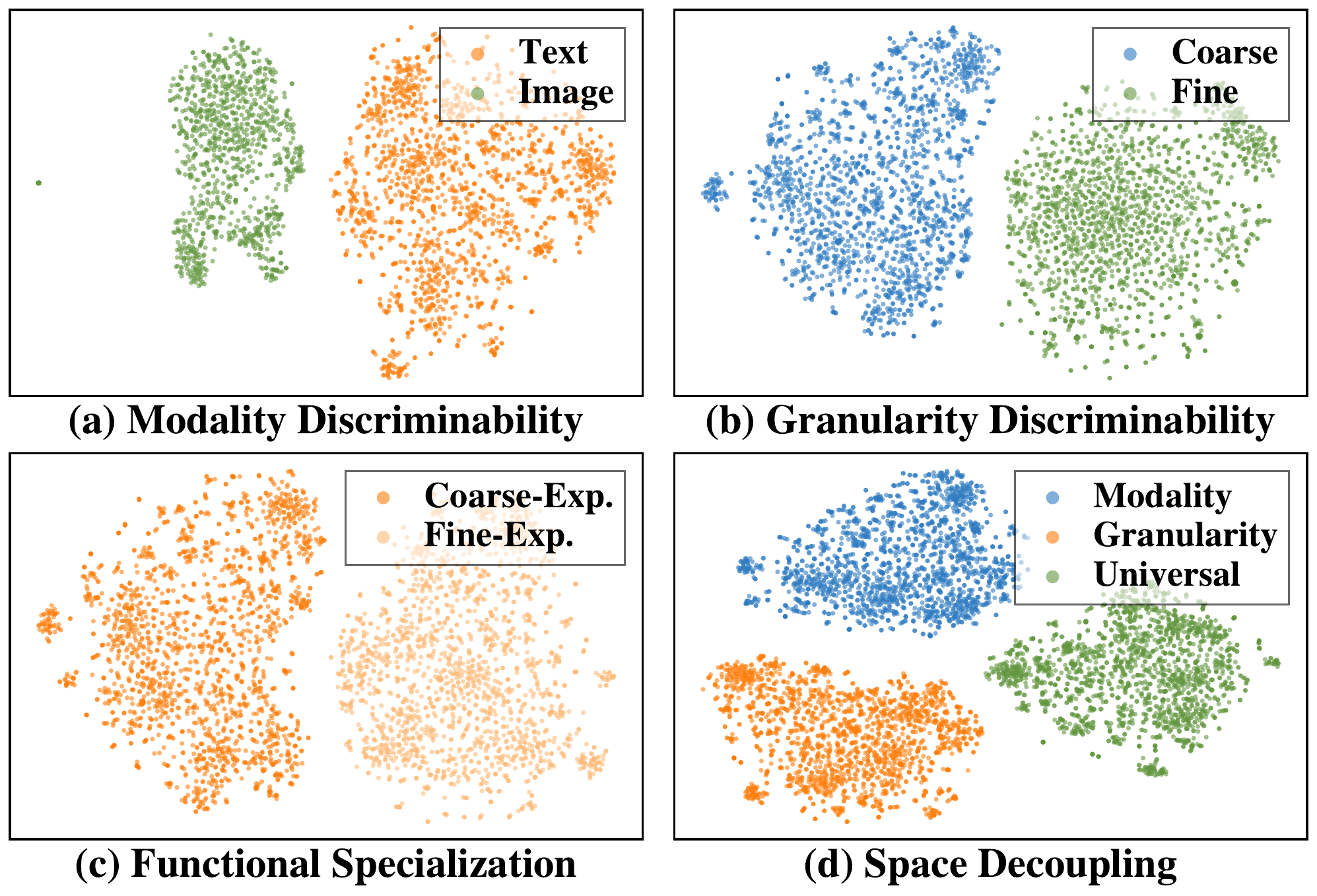}
    \caption{Different types of embedding t-SNE results on WikiDiverse dataset.}
    \label{fig7}
\end{figure}

\textbf{$\triangleright$ Effect of Temperature $\tau_{\mathcal{M}}$ and $\tau_{\mathcal{G}}$.}
We further investigate the joint impact of temperature hyperparameters $\tau_{\mathcal{M}}$ and $\tau_{\mathcal{G}}$ on model performance across the WikiDiverse and WikiMEL datasets, as illustrated in Figure \ref{fig5}. From the figure, the overall trend indicates that the model achieves optimal or near-optimal performance when both temperatures are set to moderate values (e.g., $1.0$, $0.5$). This pattern is particularly pronounced in the WikiDiverse dataset. Deviating from this range results in performance degradation, which corroborates our design rationale. Specifically, extremely low temperatures (e.g., $0.05$) induce rigid, deterministic routing that hinders necessary cross-expert knowledge sharing. Conversely, excessively high temperatures (e.g., $5.0$) lead to an overly uniform distribution—similar to a standard soft router—which weakens expert specialization and dilutes domain-specific information. based on these empirical findings, we observe a peak Hit@1 of 79.93\% on WikiDiverse when both temperatures are set to $1.0$, and 90.38\% on WikiMEL with $\tau_{\mathcal{M}}=0.5$ and $\tau_{\mathcal{G}}=0.1$.

\subsection{Further Discussion}
\hspace*{\widthof{ab}}\textbf{$\triangleright$ Analysis of Generative Semantic Refinement.}
To assess the efficacy and universality of the generative semantic refinement stage, we conducted evaluations using three distinct LLMs: DeepSeek v3 (text-only LLM), GPT-4o (multimodal LLM), and the locally deployable Qwen3-8B. As illustrated in Figure \ref{fig6}, integrating this refinement strategy yields consistent performance improvements on all baselines, and our proposed framework robustly outperforms two baseline methods (M3EL and MMOE) across all LLM backbones. Among the tested models, DeepSeek v3 demonstrates superior performance, achieving a Hit@1 of 93.55\% on WikiDiverse and 94.10\% on RichpediaMEL. Remarkably, the smaller-scale Qwen3-8B exhibits competitive performance, particularly on the RichpediaMEL dataset, where it achieves a Hit@1 of 94.08\%.

\textbf{$\triangleright$ Embedding Visualization.}
To verify the View-Specific Experts Module, we visualize expert outputs via t-SNE in Figure \ref{fig7}, corroborating our motivations from three perspectives: 
1) As shown in Figures \ref{fig7}(a) and \ref{fig7}(b), the fused outputs generated by the Modality and Granularity expert pools form distinct clusters aligned with their respective input categories (Text vs. Image; Coarse vs. Fine). Despite integrating complementary knowledge from peer experts, the aggregated representations preserve the intrinsic discriminability of the heterogeneous signals.
2) Figure \ref{fig7}(c) demonstrates that experts operate with distinct roles. When provided with identical coarse input, the Coarse (Native) and Fine (Peer) Experts still produce separable output clusters. This indicates that each expert has learned to extract unique semantic features tailored to its specific function, rather than simply replicating the input.
3) Figure \ref{fig7}(d) visualizes the outputs from the Modality, Granularity, and Universal experts given identical input. The formation of three non-overlapping clusters confirms that the Universal Expert successfully captures shared semantics, distinct from the view-specific details captured by the other experts. This disentanglement ensures a comprehensive representation that merges a stable common foundation with context-specific nuances.

\begin{figure}[t]
    \centering
    \includegraphics[width=0.9\linewidth]{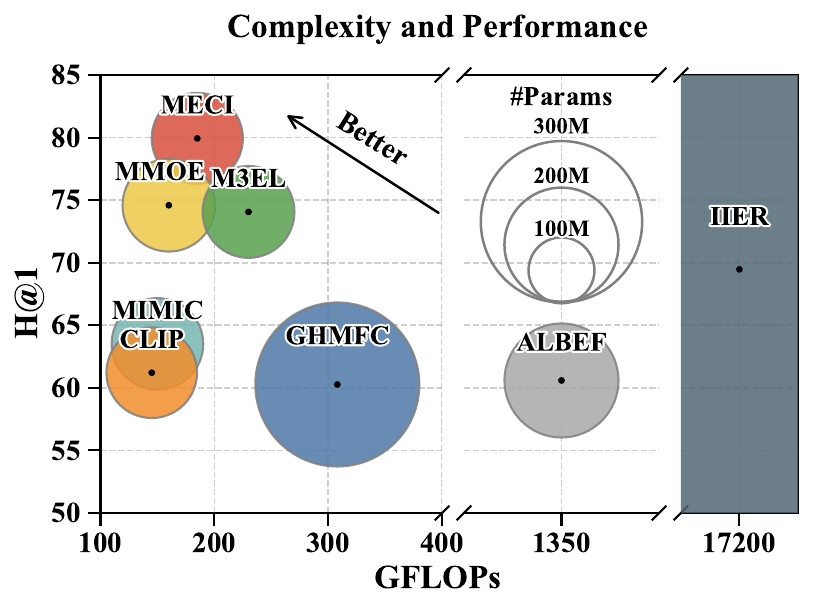}
    \caption{Comparison of model complexity and performance. The y-axis denotes the H@1 on the WikiDiverse dataset, while the x-axis represents the GFLOPs per sample. Circle size corresponds to the number of parameters.}
    \label{fig8}
\end{figure}

\textbf{$\triangleright$ Complexity Analysis.}
To evaluate the trade-off between effectiveness and efficiency, we compare the model performance (H@1) against the computational cost (GFLOPs) and model size (parameters) under the WikiDiverse dataset. As illustrated in Figure \ref{fig8}, the proposed method is located in the top-left region, which represents the better balance of high accuracy and low complexity. Existing LLM-based fine-tuning methods, such as IIER, suffer from excessive parameter counts and prohibitive computational overhead. Notably, by incorporating a training-free LLM reranking mechanism, our approach significantly outperforms the IIER method. Additionally, our model outperforms the competitive MMOE and M3EL methods in H@1 while maintaining a comparable or lower computational overhead. Furthermore, the moderate parameter size (indicated by the bubble radius) demonstrates that our approach does not rely on simply scaling up parameters to improve results. Overall, our method achieves a superior trade-off, offering a highly efficient solution for deployment.

\section{Conclusion}
We proposed the MECI framework to capture multi-level multi-element correspondences for multimodal entity linking. Specifically, to address element dominance in mention-entity correlations, we design a Multi-view Experts Network that employs a “divide-and-conquer” strategy and KL-Guided Routing to decouple feature processing. Meanwhile, to mitigate granularity-level element dominance in cross-modal correlations, we introduce a Hierarchical Multimodal Interaction Module, utilizing dynamic weighting and gating mechanisms to refine interactions. Furthermore, a generative refinement stage leverages LLMs for precise re-ranking. Experiments on WikiMEL, RichpediaMEL, and WikiDiverse confirm MECI's superiority over state-of-the-art baselines. Notably, the visual information in these datasets is sparse; thus, exploring mechanisms for missing or noisy modalities is an interesting research direction in the future.

\balance
\bibliographystyle{ACM-Reference-Format}
\bibliography{meci}

\end{document}